\documentclass[conference]{IEEEtran}
\IEEEoverridecommandlockouts
\usepackage{cite}
\usepackage{amsmath,amssymb,amsfonts}
\usepackage{bbold}
\usepackage{algorithmic}
\usepackage{graphicx}
\usepackage{textcomp}
\usepackage{xcolor}
\usepackage{tabularx}
\usepackage{tabularx}
\usepackage{booktabs}
\usepackage{stfloats}
\usepackage{siunitx}
\usepackage{placeins}
\usepackage{flushend}
\def\BibTeX{{\rm B\kern-.05em{\sc i\kern-.025em b}\kern-.08em
    T\kern-.1667em\lower.7ex\hbox{E}\kern-.125emX}}
\begin{document}

\title{Underwater Robot Pose Estimation Using\\Acoustic Methods and Intermittent Position\\Measurements at the Surface
\thanks{This work has been financially supported from H2020 SeaClear, a project that received funding from the European Union's Horizon 2020 research and innovation programme under grant agreement No 871295; by the Romanian National Authority for Scientific Research, CNCS-UEFISCDI, SeaClear support project number PN-III-P3-3.6-H2020-2020-0060; and by CNCS-UEFISCDI, project number PN-III-P2-2.1-PTE-2019-0367. We are grateful to the SeaClear team at TU Munich (Stefan Sosnoswki, Petar Bevanda, Jan Brudigam) for their development effort of the SeaClear system simulator that we use, and for initial EKF filtering results.}
}

\author{\IEEEauthorblockN{Vicu-Mihalis Maer}
\IEEEauthorblockA{\textit{Department of Automation} \\
\textit{Technical University of Cluj-Napoca}\\
Cluj-Napoca, Romania \\
vicu.maer@aut.utcluj.ro}
\and
\IEEEauthorblockN{Levente Tamás}
\IEEEauthorblockA{\textit{Department of Automation} \\
\textit{Technical University of Cluj-Napoca}\\
Cluj-Napoca, Romania \\
levente.tamas@aut.utcluj.ro}
\and
\IEEEauthorblockN{Lucian Bușoniu}
\IEEEauthorblockA{\textit{Department of Automation} \\
\textit{Technical University of Cluj-Napoca}\\
Cluj-Napoca, Romania \\
lucian.busoniu@aut.utcluj.ro}
}
\IEEEoverridecommandlockouts
\IEEEpubid{\makebox[\columnwidth]{978-1-6654-7933-2/22/\$31.00~
\copyright2022
IEEE \hfill} \hspace{\columnsep}\makebox[\columnwidth]{ }} 
\maketitle

\begin{abstract}
Global positioning systems can provide sufficient positioning accuracy for large scale robotic tasks in open environments. However, in underwater environments, these systems cannot be directly used, and measuring the position of underwater robots becomes more difficult.
In this paper we first evaluate the performance of existing pose estimation techniques for an underwater robot equipped with commonly used sensors for underwater control and pose estimation, in a simulated environment. In our case these sensors are inertial measurement units, Doppler velocity log sensors, and ultra-short baseline sensors. Secondly, for situations in which underwater estimation suffers from drift, we investigate the benefit of intermittently correcting the position using a high-precision surface-based sensor, such as regular GPS or an assisting unmanned aerial vehicle that tracks the underwater robot from above using a camera.
\end{abstract}


\section{Introduction} \label{introduction}
Underwater vehicles, whether they are remotely operated or autonomous, are widely used to execute industrial, research or military operations. Remotely operating such a vehicle can be difficult due to poor feedback. The operator has only restricted views of the scene and can also be faced with the low visibility often found in underwater environments. Also, operating multiple vehicles simultaneously in order to cooperate in executing a more complex task requires multiple operators. Consequently, Unmanned Underwater Vehicles (UUVs) are becoming more and more popular. However, in order to properly function and achieve any level of autonomy, they rely heavily on pose estimation, which is the process of estimating the position and orientation of the UUV in order to use the information as feedback for control algorithms.

In the SeaClear project\footnote{https://seaclear-project.eu}  the task we aim to fulfill is search and collection of marine litter from the bottom of water bodies using a heterogeneous multi-robot system. This requires good pose estimation for navigation purposes and litter mapping. Moreover, geo-referenced localization is required in order to enable litter items positions to be saved for later collection.

Current approaches for the underwater pose estimation problem use vision systems, acoustic systems, inertial navigation systems, or any combination of these. Conventional GPS systems cannot be directly used in underwater applications due to the water blocking the radio frequencies used.

Vision systems can use either markers as is the case in\cite{Risholm2021,Jasiobedzki2008} or existing natural features\cite{Jasiobedzki2008}. These techniques require certain conditions, such as visibility and, in the case of using natural features, their presence in the observed scene, which may not always be the case. Also, the use of markers requires preparation 
of the area, which is not possible in exploration operations.

Reference \cite{TeranEspinoza2020} uses a forward-looking sonar to compute a correction for the pose based on corresponding features from two consecutive sonar frames. Thus, they combine machine vision techniques with acoustic measurements.

Acoustic systems rely on the propagation of sound waves through water. These tend to function better than visual systems in high turbidity scenarios, where there is a higher level of light scattering and absorption. Two commonly used technologies are Doppler Velocity Logging (DVL), which measures speed relative to the bottom surface through Doppler shift, and Ultra Short Base-Line (USBL), which performs acoustic triangulation of a transceiver placed on the UUV. These are both used in\cite{Rigby2006} to obtain a position estimate.

The acoustic devices also manifest weaknesses in certain situations. For example, DVL devices are used for velocity estimation under the assumption of a fixed, rigid bottom surface, which might not always be the case due to vegetation, loose sand, etc. USBL has the great advantage that it can directly provide geo-referenced coordinates which is important to our application, but in practice the device used in our experiments has proven to not be very robust or precise, often generating outliers or sparse measurements.

In this paper, we aim to analyze the efficiency of established pose estimation techniques based on extended Kalman filters in UUVs for a litter-search and collection scenario. Our objective is to find a minimal set of sensors that can produce a satisfying pose-estimate, due to the high cost of the sensors.

We focus first on acoustic methods introduced above coupled with Inertial Measurement Units (IMUs). Secondly, we consider scenarios where underwater-only estimation suffers from large drift, e.g. due to the poor precision of either USBL or DVL. In this scenario, we investigate the idea of intermittent position measurements at the surface: the UUV periodically resurfaces and its precise position is reacquired using e.g. GPS mounted on the UUV, or a surveying Unmanned Aerial Vehicle (UAV) that locates itself using GPS and uses a camera to measure the relative displacement of the UUV. \\

Following, we will provide a description of the system including the sensors used in Section \ref{sys} and a quick introduction into Kalman filtering in Section \ref{kalman}. After this, the results obtained through fusion of the UUV on-board sensors are given in Section \ref{uuvres}. Results with intermittent surface measurements are given in Section \ref{uav} and conclusion and future prospects are presented in Section \ref{conclusion}.

\section{System description} \label{sys}
\begin{figure}[t!]
\centerline{\includegraphics[width=\linewidth]{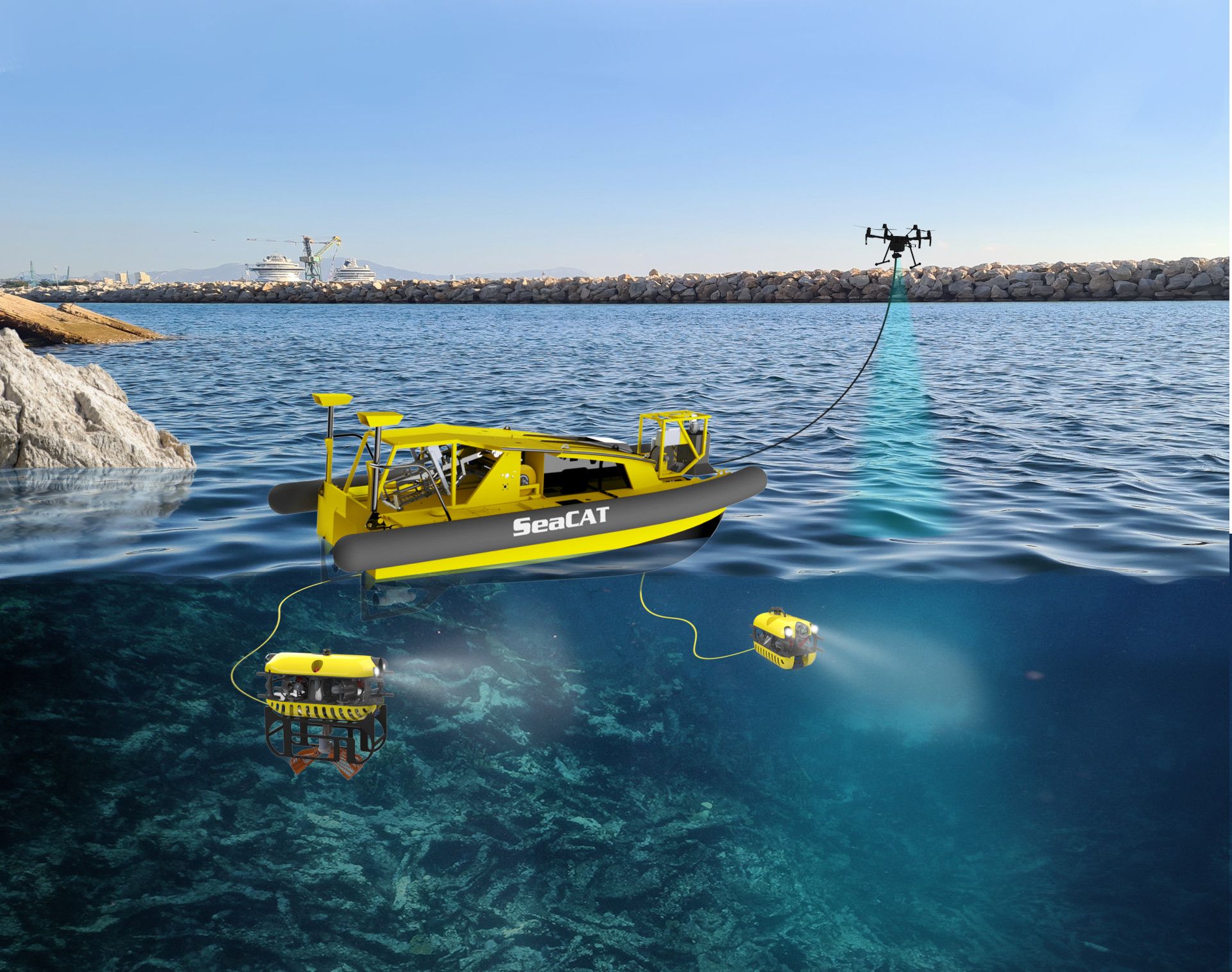}}
\caption{Robot team performing tasks.}
\label{fig:prettysystem}
\end{figure}

The SeaClear system is comprised of an autonomous surface vehicle that serves as a floating base, transporting the other robots over long distances, and two UUVs and a UAV deployed from the surface vehicle. One of the UUVs is tasked with the search of the litter while the other one collects it and deposits it in a basket attached to the surface vehicle.
Fig.~\ref{fig:prettysystem} conceptually illustrates the four-robot team.

We will now discuss the sensors mounted on the different robots, which are relevant to pose estimation.

On the UAV the key sensors are a high-precision GPS positioning system, an IMU for orientation, and a camera.

On the UUVs we have a 9 degree of freedom IMU consisting of 3 axis accelerometer, gyroscope and magnetometer, a depth-pressure sensor, a DVL which measures velocity relative to the bottom surface, and a USBL transceiver. The complementary USBL transceiver is mounted to the surface vehicle together with a GPS beacon. The GPS measurement coupled with the displacement between the two transceivers produces an absolute position of the UUV within the precision of the instruments. Data is sent from the UUV through a wired connection to a computer on the surface vehicle.

For the simulation environment, Gazebo\cite{Koenig2004} together with the Robot Operating System (ROS)\cite{Quigley} were used. For the UUVs, the \textit{uuv\_simulator} \cite{Manhaes_2016} package for ROS was used. The specific UUV model used is the SubseaTech Tortuga. In simulation the IMU and DVL sensors used are implemented in the \textit{uuv\_simulator} package. 

The IMU is modeled with zero-mean Gaussian noise and a random-walk bias. The DVL is simulated by adding a zero-mean Gaussian noise to the true velocity. The parameters for the sensors used in simulation from the package are shown in Table \ref{tab:values}. For their significance, please see\cite{Manhaes_2016}. 

The USBL measurement was simulated by adding to the ground truth zero-mean Gaussian distribution with standard deviation $\sigma=0.5$. In practice, we have found our USBL to not be very reliable, often becoming stuck and providing the same value even when the UUV is moving. We have simulated this by giving each measurement a $5\%$ chance of getting stuck and sending the same value for \SI{10}{\second}.
\begin{table}[hbtp]
    \centering
     \begin{tabular}{l l}
     \toprule
     \underline{IMU:}\\
      Gyroscope noise density & $3.394\mathrm{e}{-4}$\\
      Gyroscope random walk & $3.8785\mathrm{e}{-5}$\\
      Gyroscope bias correlation time & $1000.0$\\
      Gyroscope turn on bias sigma & $0.0087$\\
      Accelerometer noise density & $0.004$\\
      Accelerometer random walk & $0.006$\\
      Accelerometer bias correlation time & $300.0$\\
      Accelerometer turn on bias sigma & $0.1960$\\
      \underline{DVL:}\\
      Noise sigma & $0.05$\\
      Noise amplitude & $2$\\
      \underline{Pressure:}\\
      Noise sigma &$3.0$\\
      Noise amplitude & $0.0$\\
      Standard pressure & $101.325$\\
      kPaPerM & $9.80638$\\
      \bottomrule
      \vspace{0.1cm}
     \end{tabular}
     
    \caption{Parameter Values}
    \label{tab:values}
\end{table}

\section{Kalman Filters} \label{kalman}
\subsection{Basic Kalman Filter}
A Kalman filter is a linear estimator described for the first time by R.E. Kalman in 1960\cite{Kalman1960}. It is well-known and widely used in estimation problems in the area of autonomous vehicles, under various forms\cite{Welch2006}.

It aims to estimate the state of a discrete system described by the following equation:
\begin{equation}
\label{eq:statespace}
\begin{aligned}
  \mathbb{x}_k &= A\cdot\mathbb{x}_{k-1} + B\cdot \mathbb{u}_{k-1} + w_k\\
  \mathbb{y}_k &= H\cdot \mathbb{x}_k + v_k
\end{aligned}
\end{equation}
where:

     \begin{tabular}{c l}
      $\mathbb{x}_k$ & State vector at step k\\
      $A, B, H$ & State-space matrices\\
      $\mathbb{y}_k$ & Measurement at step k\\
      $w_k$ & Process noise\\
      $v_k$ & Measurement noise\\
     \end{tabular}
   
The noise, $w_k$ and $v_k$ are considered to be normally distributed with zero mean and covariance matrices $Q$ and $R$ respectively.

The estimation works in two steps: prediction and correction. For the prediction step, the state at step $k+1$ is computed using the model as shown in \eqref{eq:prediction}. Also an estimate error covariance $P$ is computed.
\begin{equation}
\label{eq:prediction}
\begin{aligned}
  \hat{\mathbb{x}}_{k} &= A\cdot\hat{\mathbb{x}}_{k-1} + B\cdot \mathbb{u}_{k-1}\\
  P_{k} &= AP_{k-1}A^\top + Q
\end{aligned}
\end{equation}
After the prediction step, a correction is applied based using the measurement.
\begin{equation}
\label{eq:correction}
\begin{aligned}
    K_k &= P_kH_k^\top(H_kP_kH_k^\top+R)^{-1}\\
      \hat{\mathbb{x}}_{k} &= \hat{\mathbb{x}}_k + K_k( \mathbb{y}_k-H\hat{\mathbb{x}}_k)\\
  P_{k} &= (I-K_kH)P_k
\end{aligned}
\end{equation}

When using the filter, the covariance matrices $Q$ and $R$ play a big role in the accuracy of the estimate, so they must represent the process and measurement noise as accurately as possible. Often they are empirically tuned.
\subsection{Extended Kalman Filter}
The Extended Kalman Filter aims to solve the estimation problem when the system whose states are being estimated is non-linear.
It does this by linearizing the system at each step in order to compute the correction.
In the nonlinear case the system is defined by the following equations
\begin{equation}
\label{eq:nonlinear}
\begin{aligned}
  \mathbb{x}_k &= f(\mathbb{x}_{k-1}, \mathbb{u}_{k-1}, w_k)\\
  \mathbb{y}_k &= h(\mathbb{x}_{k}, v_k)
\end{aligned}
\end{equation}
where $f$ and $h$ are non-linear functions that characterize the evolution of the states and the measurement respectively.
When predicting, the noise will be assumed zero.
After linearization, the equations become:
\begin{equation}
\label{eq:linearized}
\begin{aligned}
  \mathbb{x}_k &\approx f(\hat{\mathbb{x}}_{k-1}, \mathbb{u}_{k-1}, 0) + A_k\cdot(\mathbb{x}_{k-1}-\hat{\mathbb{x}}_{k-1}) + W_k\cdot w_k\\
  \mathbb{y}_k &\approx h(\hat{\mathbb{x}}_{k}, 0) + H_k\cdot(\mathbb{x}_{k-1}-f(\hat{\mathbb{x}}_{k-1}, \mathbb{u}_{k-1}, 0)) + V_k\cdot v_k
\end{aligned}
\end{equation}
where $A_k$, $H_k$, $W_k$ and $V_k$ are Jacobian derivative matrices of functions $f$ and $h$ with respect to $\mathbb{x}_k$, $w_k$ and $v_k$

In this case, the prediction and correction equations become:
\begin{equation}
\label{eq:ekf}
\begin{aligned}
  \hat{\mathbb{x}}_{k+1} &= f(\hat{\mathbb{x}}_{k-1}, \mathbb{u}_{k-1}, 0)\\
  P_{k} &= A_{k-1}P_{k-1}A_{k-1}^\top + W_{k-1}Q_{k-1}W_{k-1}^\top\\
   K_k &= P_kH_k^\top(H_kP_kH_k^\top+V_kR_kV_k^\top)^{-1}\\
      \hat{\mathbb{x}}_{k} &= \hat{\mathbb{x}}_k + H_k( \mathbb{y}_k-h(\hat{\mathbb{x}}_{k}, 0))\\
  P_{k} &= (I-K_kH_k)P_k
\end{aligned}
\end{equation}

For our purposes, the \textit{robot\_localization} package, described in\cite{Moore2016}, was used for the implementation of the Extended Kalman Filter algorithm. It is a software package for ROS that includes, among others, an implementation of the Extended Kalman Filter ready to be used for pose estimation using various sensor types.
\begin{figure}[t!]
\centerline{\includegraphics[width=0.75\linewidth]{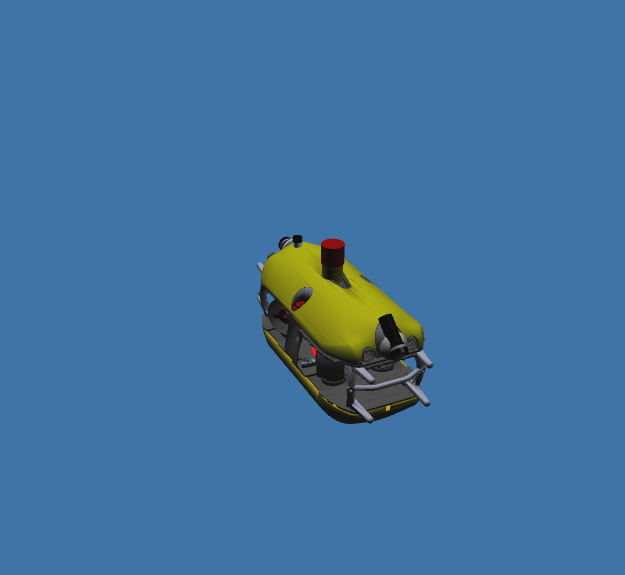}}
\caption{UUV in simulation environment}
\label{fig:uuvscreenshot}
\end{figure}

For the  $f$ function from \eqref{eq:nonlinear}, the \textit{robot\_localization} uses a standard Newtonian 3D rigid-body kinematic model \cite{Moore2016}. As for the measurement function $h$, it is assumed that each sensor measures the states directly, with additive Gaussian noise\cite{Moore2016}.

In our experiments below, the filter was run with a sampling rate of \SI{20}{\hertz}. The process noise covariance matrix used is given in the following equation.
\begin{equation}
\label{eq:processNoise}
\begin{aligned}
  Q = \mathrm{diag}(\left[\begin{matrix} 1\mathrm{e}{-3} & 1\mathrm{e}{-3} & 1\mathrm{e}{-3} & 0.3 & 0.3 & 0.3 & 0.5\end{matrix}\right.\\ 
    \left.\begin{matrix}& 0.5 & 0.1 & 0.3 & 0.3 & 0.3 & 0.3 & 0.3 & 0.3\end{matrix}\right] )
\end{aligned}
\end{equation}

The noise covariance for the sensor measurements is equal to the covariances of the sensors configured in the simulator and given in Table \ref{tab:values}.

\section{Underwater-Only Estimation Performance}\label{uuvres}
In the first batch of experiments, we have performed simulations using only the sensors available underwater in different configurations. In all cases, a pressure based depth sensor was used for the vertical axis position. Here are the different configurations which will be discussed individually:
\begin{itemize}
    \item IMU only
    \item IMU+USBL
    \item IMU+DVL
    \item IMU+DVL+USBL
\end{itemize}

Table \ref{tab:mse} compares the mean squared error (MSE) for the estimated pose with all these configurations.

\begin{table*}[!t]
\caption{MSE for underwater-only pose estimation}
\begin{center}
\begin{tabular}{|c|c|c|c|c|c|c|}
\hline
\textbf{Table}&\multicolumn{6}{|c|}{\textbf{Axes}} \\
\cline{2-7} 
\textbf{Sensor Configuration} & \textbf{\textit{X(m)}}& \textbf{\textit{Y(m)}}& \textbf{\textit{Z(m)}}& \textbf{\textit{Roll(rad)}}& \textbf{\textit{Pitch(rad)}}& \textbf{\textit{Yaw(rad)}} \\
\hline
IMU & 18.384 & 15.963 & 0.031155 & 9.72e-09 & 9.1397e-09 & 0.0090529\\
\hline
IMU+USBL & 2.4513 & 1.1686 & 0.031155 & 9.9215e-09 & 9.2075e-09 & 0.014226\\
\hline 
IMU+DVL & 0.041847 & 0.14544 & 0.03108 & 9.9407e-09 & 9.202e-09 & 0.0090492\\
\hline
IMU+DVL+USBL & 0.93606 & 0.28376 & 0.03108 & 9.888e-09 & 9.3111e-09 & 0.01228 \\
\hline

\end{tabular}
\label{tab:mse}
\end{center}
\end{table*}

The UUV was manually driven on a rectangular trajectory once during the experiments returning close to the starting point at the end. Fig. \ref{fig:uuvscreenshot} shows a capture from the simulation environment.

\subsection{IMU only}
\begin{figure}
\centerline{\includegraphics[width=\linewidth]{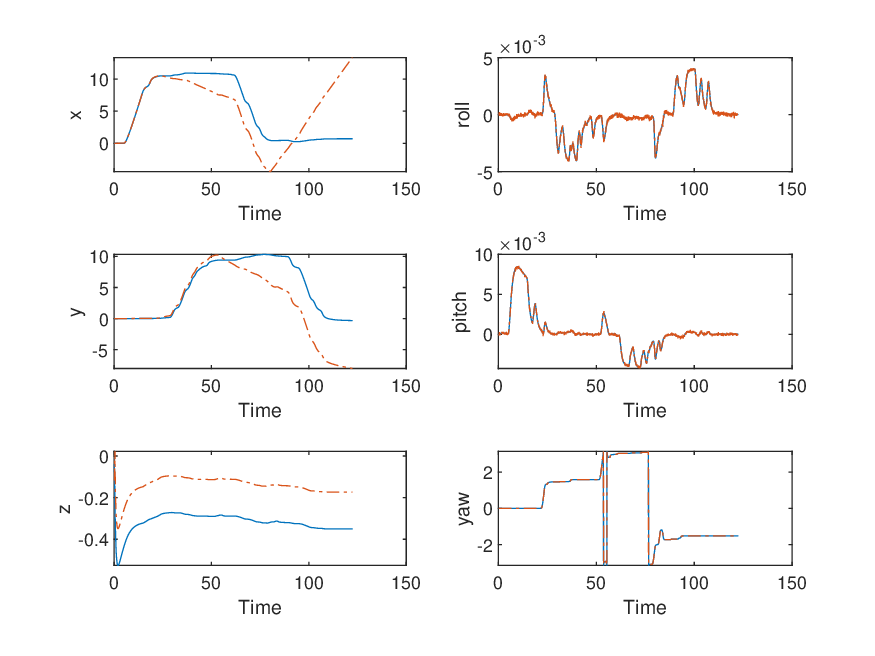}}
\caption{Pose estimate (orange) and ground truth (blue) for IMU only sensor configuration}
\label{fig:imu}
\end{figure}

Fig. \ref{fig:imu} shows the pose-estimate compared to the ground truth. As expected, the values drift making the use of such an estimate reliable only for short periods of time. Even though the other devices might produce more accurate measurements, the IMU is essential for orientation estimation.

\subsection{IMU+DVL}
\begin{figure}[t!]
\centerline{\includegraphics[width=\linewidth]{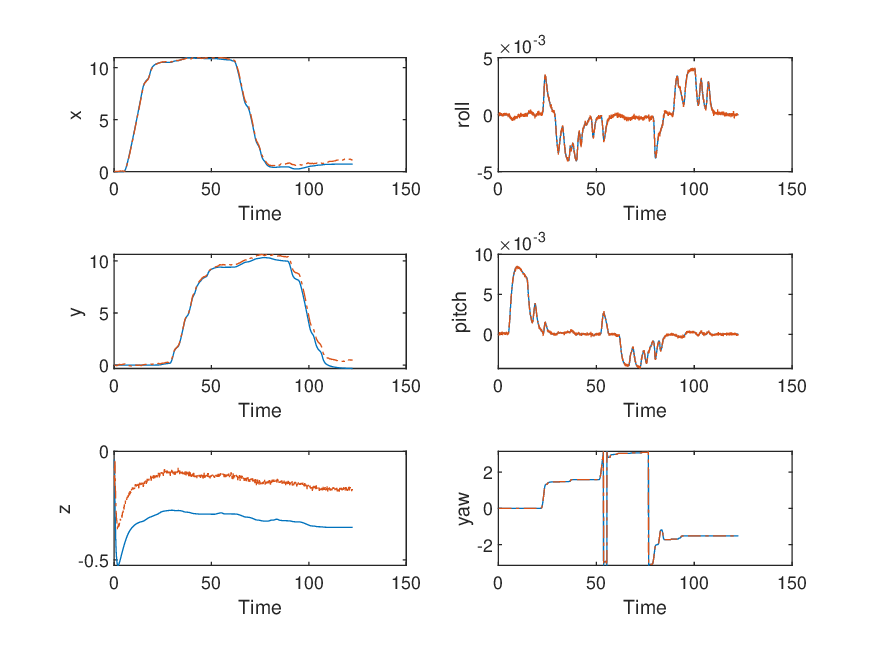}}
\caption{Pose estimate (orange) and ground truth (blue) for IMU and DVL sensor configuration}
\label{fig:dvl}
\end{figure}

The estimation results are shown in Fig. \ref{fig:dvl}. Due to the single integration step, less drift is expected when using velocity measurements. This is confirmed by the results obtained where a smaller amount of drift is present by the end of the experiment.
\subsection{IMU+USBL}
\begin{figure}[t!]
\centerline{\includegraphics[width=\linewidth]{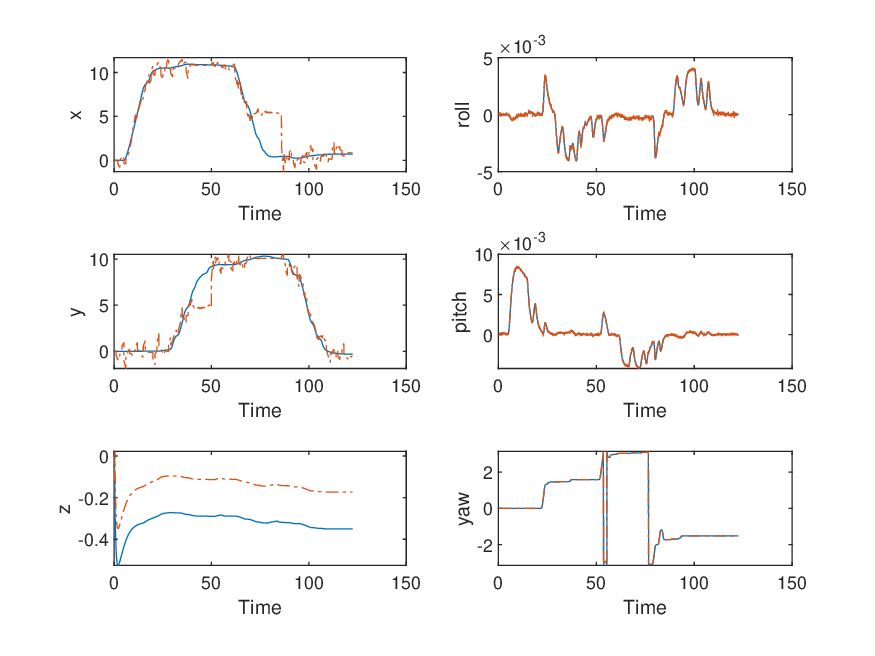}}
\caption{Pose estimate (orange) and ground truth (blue) for IMU and USBL sensor configuration}
\label{fig:usbl}
\end{figure}

Fig. \ref{fig:usbl} shows the pose-estimate when using IMU and USBL together. We can observe that the measurements do not drift, even though they are noisy. Also, the USBL satisfies the need of a geo-referenced estimate. If high accuracy position is not required, an IMU-USBL pair can be sufficient in order to obtain a 6-DOF estimate.
\subsection{IMU+DVL+USBL}
The error for this sensor configuration shows that the DVL improves the accuracy of the estimate. Another important aspect, although not visible in this instance due to the short length of the experiment, is that the USBL will prevent the drift caused by integrating noisy DVL measurements.
\begin{figure}[hbp]
\centerline{\includegraphics[width=\linewidth]{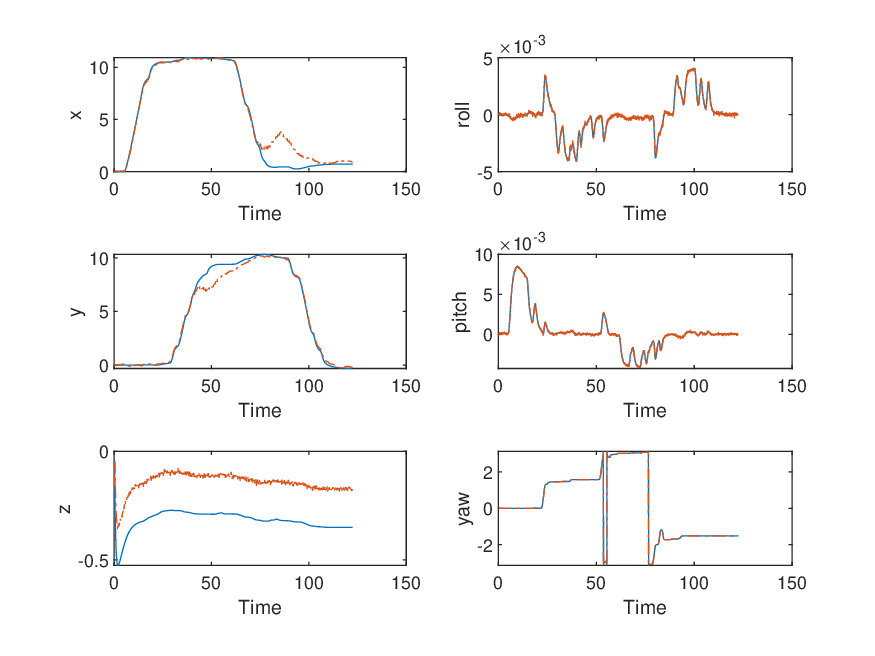}}
\caption{Pose estimate (orange) and ground truth (blue) for IMU, DVL and USBL sensor configuration}
\label{fig:usbl_dvl}
\end{figure}

\section{Surface-based intermittent positioning}\label{uav}

In the following experiments we remove the DVL from the UUV due to its high cost. In this USBL+IMU configuration with a poor USBL, the UUV suffers from significant position drift. 

One alternative to mitigate this drift is to resurface the UUV periodically to reacquire a high-precision position estimate with an additional sensor. This could be a waterproofed GPS sensor mounted on the UUV. A more interesting alternative is given by the observation that in sufficiently clear waters or shallow depths, the UUV is visible through the UAV's camera. In such moments, knowing the GPS coordinates of the UAV, the position of the UUV in the camera image and its depth can be used to compute a more precise estimate for the position of the UUV. 

Our experiment generically simulates any such surface sensor by corrupting the ground truth with Gaussian noise with $\sigma = 0.05$ (corresponding to the data-sheet accuracy of the GPS device used on the UAV).

\begin{figure}[t!]
\centerline{\includegraphics[width=\linewidth]{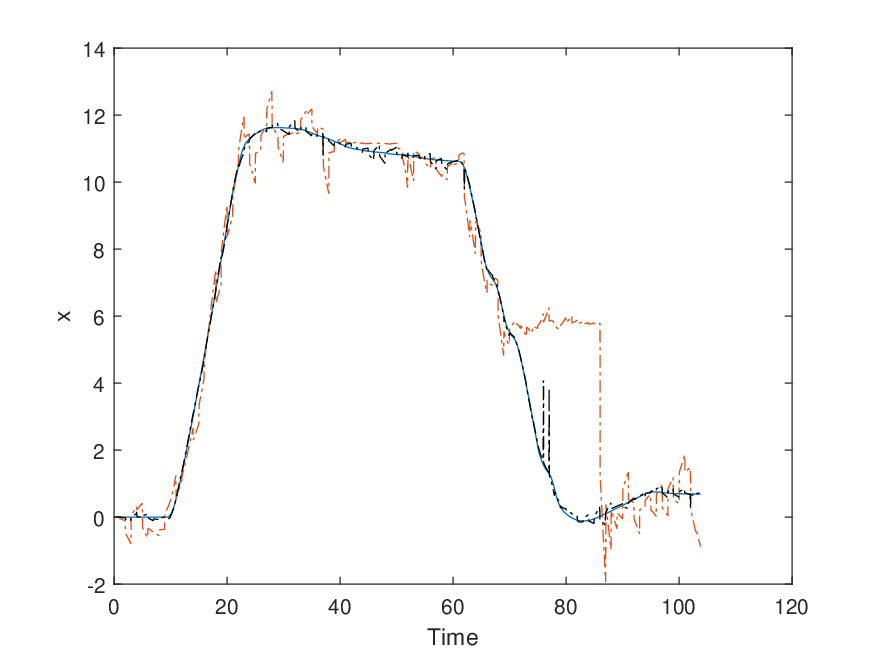}}
\caption{Ground truth (blue), pose estimate without surface feedback (orange) and pose estimate with surface feedback (black).}
\label{fig:usbl_uav}
\end{figure}

The experiments were performed with three different feedback (``resurfacing'') periods. Although these periods are quite small and practically correspond to the situation in which the UUV is constantly visible, choosing higher periods would not make sense due to the short experiment duration. Fig. \ref{fig:usbl_uav} shows results. Here, the UUV's position on one of the axes for the 1 sec feedback period is shown alongside the scenario with only USBL and no feedback. We can see that the surface feedback brings the estimate closer to the ground truth.

Table \ref{tab:uavmse} shows the MSE obtained in 4 different scenarios using the IMU, USBL and feedback from the surface. The angles have been omitted since the surface sensors do not give any information on the orientation of the UUV. 
As expected, a lower update period gives lower errors, however these periods have to be kept quite low due to the noisiness of the USBL data. This scenario corresponds to the UUV only spending time at the surface.

Seeing that, in IMU only experiments, the IMU only estimate is stable for periods longer than the surface measurement period, we also tried the scenario in which the UUV is equipped only with an IMU and receives surface measurements every \SI{30}{\second}.

\begin{figure}[t!]
\centerline{\includegraphics[width=\linewidth]{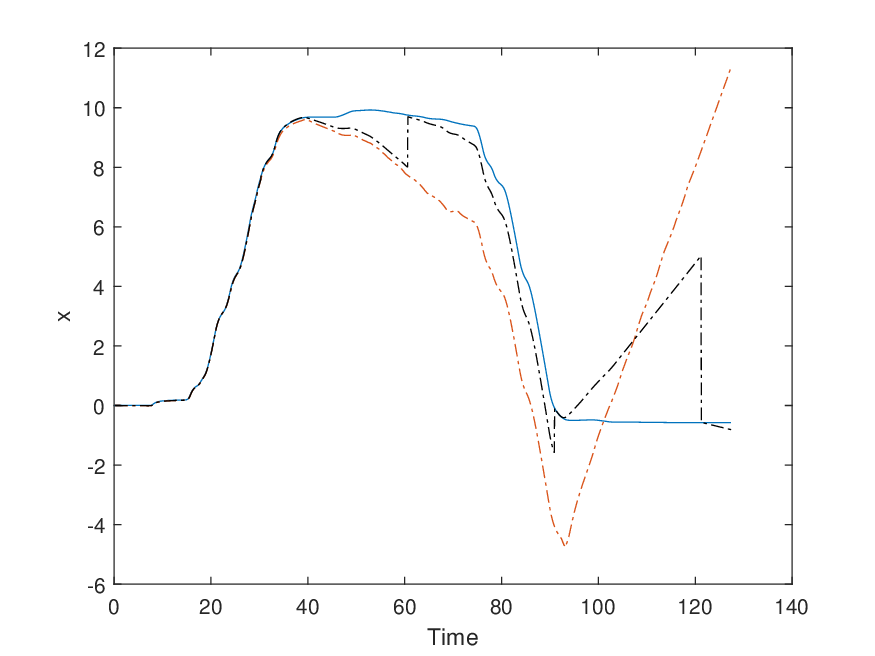}}
\caption{Ground truth (blue), pose estimate without surface feedback (orange) and pose estimate with surface feedback (black).}
\label{fig:imu_uav}
\end{figure}

The results are shown in Fig. \ref{fig:imu_uav} again, only for one axis, for clarity. The MSE for the position is shown in Table \ref{tab:imuuavmse}. In this scenario, we obtain better estimates and \SI{30}{\second} is a more realistic update period, but we eill still have larger drifts sometimes as seen around \SI{100}{\second}.

\begin{table}[htbp]
\caption{Estimate MSE for USBL with surface position measurements}
\begin{center}
\begin{tabular}{|c|c|c|c|}
\hline
\textbf{Table}&\multicolumn{3}{|c|}{\textbf{Axes}} \\
\cline{2-4} 
\textbf{Surface feedback period} & \textbf{\textit{X(m)}}& \textbf{\textit{Y(m)}}& \textbf{\textit{Z(m)}} \\
\hline
no feedback & 3.5328 & 2.3847 & 0.0311469\\
\hline
1 sec & 0.01391 & 0.018928 & 0.031146 \\
\hline 
5 sec & 3.0389 & 2.466 & 0.031146 \\
\hline
10 sec &3.3012 & 2.5037 & 0.031146 \\
\hline

\end{tabular}
\label{tab:uavmse}
\end{center}
\end{table}

\begin{table}[htbp]
\caption{Estimate MSE for IMU with surface position measurements}
\begin{center}
\begin{tabular}{|c|c|c|c|}
\hline
\textbf{Table}&\multicolumn{3}{|c|}{\textbf{Axes}} \\
\cline{2-4} 
\textbf{Surface feedback period} & \textbf{\textit{X(m)}}& \textbf{\textit{Y(m)}}& \textbf{\textit{Z(m)}} \\
\hline
IMU only & 13.236 & 18.773 & 0.031273\\
\hline
30 sec & 2.5694 & 0.96057 & 0.031273 \\
\hline
\end{tabular}
\label{tab:imuuavmse}
\end{center}
\end{table}

\section{Conclusion}\label{conclusion}

We performed simulations in order to asses the accuracy of Extended Kalman filtering for pose estimate of unmanned underwater robots, using different sensor configurations. Acoustic and inertial sensors have been used for the intrinsic pose estimation of the UUV. For the situation in which underwater-only estimation does not work well, periodic position corrections using a surface-based sensor were proposed and evaluated.

Further experiments using the mentioned sensor configurations are planned both in simulated environments and using real hardware. Also, various extensions brought to the Kalman filter such as the unscented Kalman filter, could be tested in the future along additional analysis of the estimate under intermittent feedback.


\vspace{12pt}
\bibliographystyle{ieeetr}
\bibliography{bibl}

\begin{thebibliography}{10}

\bibitem{Risholm2021}
P.~Risholm, P.~{Ornulf Ivarsen}, K.~{Henrik Haugholt}, and A.~Mohammed,
  ``{Underwater marker-based pose-estimation with associated uncertainty},'' in
  {\em Proceedings of the IEEE International Conference on Computer Vision},
  vol.~2021-Octob, pp.~3706--3714, 2021.

\bibitem{Jasiobedzki2008}
P.~Jasiobedzki, S.~Se, M.~Bondy, and R.~Jakola, ``{Underwater 3d mapping and
  pose estimation for ROV operations},'' in {\em OCEANS 2008}, 2008.

\bibitem{TeranEspinoza2020}
A.~{Teran Espinoza}, {\em {Acoustic-Inertial Forward-Scan Sonar Simultaneous
  Localization and Mapping}}.
\newblock PhD thesis, 2020.

\bibitem{Rigby2006}
P.~Rigby, O.~Pizarro, and S.~B. Williams, ``{Towards geo-referenced AUV
  navigation through fusion of USBL and DVL measurements},'' in {\em OCEANS
  2006}, 2006.

\bibitem{Koenig2004}
N.~Koenig and A.~Howard, ``Design and use paradigms for gazebo, an open-source
  multi-robot simulator,'' {\em 2004 IEEE/RSJ International Conference on
  Intelligent Robots and Systems (IROS)}, vol.~3, pp.~2149--2154, 2004.

\bibitem{Quigley}
M.~Quigley, K.~Conley, B.~Gerkey, J.~Faust, T.~Foote, J.~Leibs, R.~Wheeler, and
  A.~Ng, ``Ros: an open-source robot operating system,'' in {\em Proc. of the
  IEEE Intl. Conf. on Robotics and Automation (ICRA) Workshop on Open Source
  Robotics}, 2009.

\bibitem{Manhaes_2016}
M.~M.~M. Manh{\~{a}}es, S.~A. Scherer, M.~Voss, L.~R. Douat, and
  T.~Rauschenbach, ``{UUV} simulator: A gazebo-based package for underwater
  intervention and multi-robot simulation,'' in {\em {OCEANS} 2016 {MTS}/{IEEE}
  Monterey}, {IEEE}, sep 2016.

\bibitem{Kalman1960}
R.~E. Kalman, ``{A new approach to linear filtering and prediction problems},''
  {\em Journal of Fluids Engineering, Transactions of the ASME}, vol.~82,
  no.~1, pp.~35--45, 1960.

\bibitem{Welch2006}
G.~Welch and G.~Bishop, ``{An Introduction to the Kalman Filter},'' {\em In
  Practice}, vol.~7, no.~1, pp.~1--16, 2006.

\bibitem{Moore2016}
T.~Moore and D.~Stouch, ``{A generalized extended Kalman filter implementation
  for the robot operating system},'' in {\em Advances in Intelligent Systems
  and Computing}, vol.~302, pp.~335--348, 2016.

\end{thebibliography}

\end{document}